\documentclass[conference]{IEEEtran}

\usepackage{hyperref}
\usepackage{cite}
\usepackage{amsmath,amssymb,amsfonts}
\usepackage{algorithmic}
\usepackage{graphicx}
\usepackage{textcomp}
\usepackage{xcolor}
\usepackage{siunitx}

\def\BibTeX{{\rm B\kern-.05em{\sc i\kern-.025em b}\kern-.08em
    T\kern-.1667em\lower.7ex\hbox{E}\kern-.125emX}}
\begin{document}

\title{Systematically designing better instance counting models on cell images with Neural Arithmetic Logic Units}

\author{\IEEEauthorblockN{Ashish Rana\IEEEauthorrefmark{1},
Taranveer Singh\IEEEauthorrefmark{2}, Harpreet Singh\IEEEauthorrefmark{3}, Neeraj Kumar\IEEEauthorrefmark{4} and Prashant Singh Rana\IEEEauthorrefmark{5}}
\IEEEauthorblockA{Department of Computer Science,
TIET Patiala, Punjab, India.\\
Email: \IEEEauthorrefmark{1}\{arana\_be15, tsingh\_me17, harpreet.s, neeraj.kumar and prashant.singh \}@thapar.edu
}}

\maketitle

\begin{abstract}

The big problem for neural network models which are trained to count instances is that whenever testing range data goes to higher counts than training range data generalization error increases i.e. prediction error for images that are outside training range increases. Consider the case of automating cell counting process where more dense images with higher cell counts are commonly encountered as compared to images used in training data.
By making better predictions for higher ranges of cell count we are aiming to create better generalization systems for cell counting. With architecture proposal of neural arithmetic logic units (NALU) for arithmetic operations, task of counting has become feasible for higher numeric ranges which were not included in training data with better accuracy. In our study we incorporate these units in already existing Fully Convolutional Regression Network (FCRN) and U-Net architectures in the form of residual concatenated layers. We carried out a systematic comparative study with the newly proposed changes and earlier base architectures. This comparative study results are evaluated in terms of optimizing regression loss across cell density map over an image. We achieved better results in cell counting tasks with our newly proposed architectures having residual layer concatenation connections. We further validated our results on custom created high count dataset created from BBBC005 synthetic cell count dataset and obtained even better results where testing images have higher counts. These results confirm that above mentioned numerically biased units does help models to learn numeric quantities for better generalization results on high count data.

\iffalse
Our experiments on a batch size of 16 shows a 20.22\% relative improvement as compared to original architecture and 34.84\% relative improvement on our custom created high count dataset created from BBBC005 synthetic cell count dataset achieving higher generalization capabilities for U-net based architectures.
\fi

\begin{IEEEkeywords}
Neural Arithmetic Logic Units, Cell Counting, Fully Convolutional Regression Networks, U-net.
\end{IEEEkeywords}

\end{abstract}

\section{Introduction}

Ability to generalize concepts is fundamental component of intelligence and core for designing
smart systems \cite{b1, b2}. Neural networks simulates this behavior with hierarchical learning of concepts. When it comes to automation, counting is an important task from machine vision application \cite{b3} to cell counting \cite{b4}. While neural networks manipulates numerical quantities but it is not associated with systematic generalization \cite{b5, b6}. These networks fail to generalize as evident from high generalization error while predicting quantities that lie outside the training numerical range \cite{b7}. This highlights memorization behavior in neural networks instead of generalization abilities for a given task. This is especially problematic for cell counting tasks as images with higher cell counts that were not part of training are common to encounter in real applications.

Neural accumulators (NAC) and neural arithmetic logic units (NALU) \cite{b7} are biased to learn
systematic numerical computation and performs relatively better than non linear activation functions for arithmetic operations. This numerical bias of learning computations makes
them excellent choice for counting tasks which are essentially is an increment addition operation
only. Deep learning models generally take either segmentation approach with explicit counting trainer or end-to-end counting via a regression loss. In this paper we will go through the
latter approach \cite{b4} in detail for automation of cell counting process. As cell counting is cumbersome task and dense cell images with higher cell counts containing data outside training numeric range are common in real world scenarios. Achieving true cell automation with less generalization errors is the prime objective of this paper.

\begin{figure}[!h]
\centering
\includegraphics[width=0.47\textwidth]{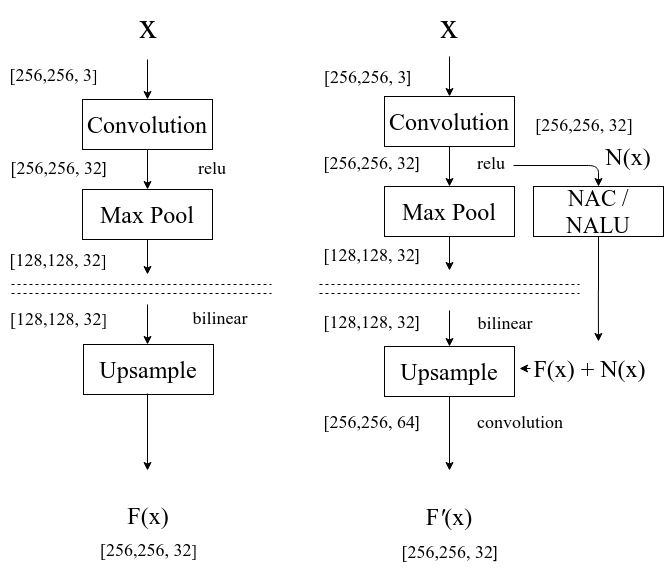}
\caption{Abstract representation of proposed modified architecture (\textit{Right}) with either NAC or NALU as compared to previous Fully Convolution Regression Networks (\textit{Left}). Here, as clearly specified with dimensional analysis of tensor blocks (\textit{Right}) the layers in the newly proposed model are concatenated and again squeezed into same size as earlier base model architecture.}
\label{fig1}
\end{figure}

In regression loss approach Fully convolutional regression networks \cite{b2} and U-net \cite{b8} architectures learns mapping  between an image I(x) and a density map D(x), given by F$\colon$ I(x) $\rightarrow$ D(x) (I $\in$ R\textsuperscript{m $\times$ n} , D $\in$ R\textsuperscript{m $\times$ n} ) for a m $\times$ n pixel image. Later on, variations of these base architectures with different activation functions are implemented in this paper and their prediction performance is compared with newly proposed NAC/NALU concatenated architectures of FCRN and U-net. The concatenated connections adds numerical bias to layers of the network and behaves in similar manner like ResNets \cite{b9}. But, instead of addition of inputs from previous layer the input from previous is passed through these numerical units for capturing numerical bias and concatenated with the main network layer as shown in figure \ref{fig1}. On surface it appears that these proposed architectural changes leads to accuracy improvements due to increased model capacity with these numerically biased units adding more parameters for learning. But, our results with custom high count dataset for testing created from BBBC005 \cite{b10} reflects increased generalization counting abilities. As for higher cell count images our model gives higher relative improvement in mean absolute error(MAE).

Our concatenation based residual architecture utilizes the fundamentals of batch normalization like specified identity mapping architecture \cite{b11} in ResNets. But, instead of using convolution operation directly this network leverages numerical bias information obtained from NAC and NALU operations applied on input layer and then finally uses convolution operation on the concatenated layer. Before and after this concatenation of this numerical bias learning operation, batch normalization is carried out and output of this operation is added back again to our next main network layer, as shown in figure  \ref{fig1}.

With means of this paper we introduce changes in current regression based model architectures for end-to-end counter training and produce systems with improved accuracy. Also, we validate our trained models on a different specially tailored validation dataset with approximately seven times higher counts of cells as compared to training dataset created from BBBC005 synthetic cell dataset \cite{b10}. Overall, experimental results demonstrates that supplementing the base architectures with NAC and NALU helped in achieving better results and improved relative MAE for higher count test images. 

\section{Related Work}

Intuitive numerical understanding is important in learning and by adjunct important in deep learning \cite{b1} for creating better models with higher generalization capabilities. Counting objects \cite{b12, b13, b14, b15, b16} in given image is a widely studied task. Trained models for counting tasks either use a deep learning model to segment instances of given object then count them in a post-processing step \cite{b17} or learn end-to-end predict count via a regression loss \cite{b4}.  Networks like Count-ception \cite{b18} added the concept of average over redundant predictions with its deep Inception family network.  Also, recent architectures like ResNets \cite{b19}, Highway Networks \cite{b20} and Densenets \cite{b21} advocate linear connections like Count-ception to promote better learning bias. Such models have better performance, though additional computational overhead due to increased depth of given architectures do arise. Our work highlights the generalization capabilities of the network, that extrapolate well on unseen parts of solution space which highlights underlying structure of behavior governing-equations \cite{b22}.

 We introduce architectural changes in models that learns and preserves input information which is numerically biased with reference to input layers. It is somewhat similar to ResNets \cite{b9}, which are easier to optimize and gain accuracy with increasing depth. With our experiments we aim to highlight that models with numerically biased concatenated residual functions helps in achieving better results with their addition in the form of a comparative study with original architectures. Also, with our results in this paper we demonstrate with our results that backpropagation learns this numerical bias without any explicit numeric quantity being provided as input implying that better computer vision counters can be trained with this module when added to existing convolutional neural network architectures.

Density based estimation doesn’t require prior object detection or segmentation \cite{b15, b12, b23}. In previous years, several works have investigated this approach. In \cite{b15}, the problem is stated as density estimation with a supervised learning algorithm, D(x) = c\textsuperscript{T}$\phi$(x), where D(x) represents  ground-truth density map, and $\phi$(x) represents  local features and parameters \textit{c} are learned by minimizing the error between predicted and true density with quadratic programming over all possible sub-windows. In \cite{b23}, regression forest is used to exploit patch-based idea for learning structured labels, then for new input image density map is estimated averaged over structured patch-based predictions. Also, in \cite{b12} an algorithm is used that allows fast interactive counting with ridge regression. 

\begin{figure}[!h]
\centering
\includegraphics[width=0.50\textwidth]{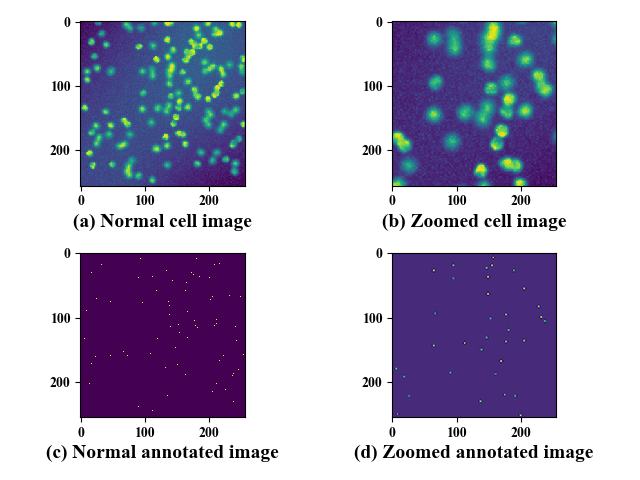}
\caption{The left column represents training image in first row and corresponding annotated image in second row from fluorescent synthetic dataset \cite{b29}. The second column of highlights corresponding zoomed view for better visual comprehension of the images in the dataset.}
\label{fig2}
\end{figure}

Cell counting \cite{b15} problem is classified into supervised learning problem that learns mapping between an image I(x) and a density map D(x), denoted by F$\colon$ I(x) $\rightarrow$ D(x) (I $\in$ R\textsuperscript{m $\times$ n} , D $\in$ R\textsuperscript{m $\times$ n} ) for a m $\times$ n pixel image, see figure \ref{fig2}. Density function D(x) function is defined on pixels in given image, integrating this map over an image region gives an estimate of number of cells in that region.  CNNs \cite{b24, b25} are quite popular in the bio-medical imaging because of their simple architecture and achieve great results. Like in mitosis detection \cite{b26}, neuronal membrane segmentation \cite{b27} and analysis of C. elegans embryos development \cite{b28}. Previously, fully convolutional regression networks (FCRNs) and Count-ception have given state-of-the-art results in cell counting, with potential for cell detection of overlapping cells.

Also U-Nets \cite{b8}, typically fully convolutional network uses a modified version of architecture proposed by Ciresan et al. \cite{b27} as latter is slow and trade-off between localization and use of context are present. In U-Nets pooling operations are replaced by upsampling operations to supplement usual contracting network. For localization high resolution features from contracting path are combined with unsampled output. Based on this information a successive convolution layer then learn to assemble more precise output. For our experimentation we selected FCRN and U-net based on simplicity and relative similarity in their architectures with the difference being that U-net uses inputs from previous layer for better localization.

\section{Experiments}

In this section first we conceptually explore NACs and NALUs. Then compare their addition capabilities with multi-layer perceptrons equipped with different activation functions. With this study we aim to select NAC/NALU designed variants which can best approximate the counting behavior and compare them with standard FCRN and U-net neural network architecture for regression loss approach in the following experiment done on synthetic dataset \cite{b29}. For validation of counting generalization capabilities, our trained models are tested against different synthetic cell image dataset with approximately seven times higher counts than training data.

\subsection{Visual understanding of NACs and NALUs}

\begin{figure}[!h]
\centering
\includegraphics[width=0.55\textwidth]{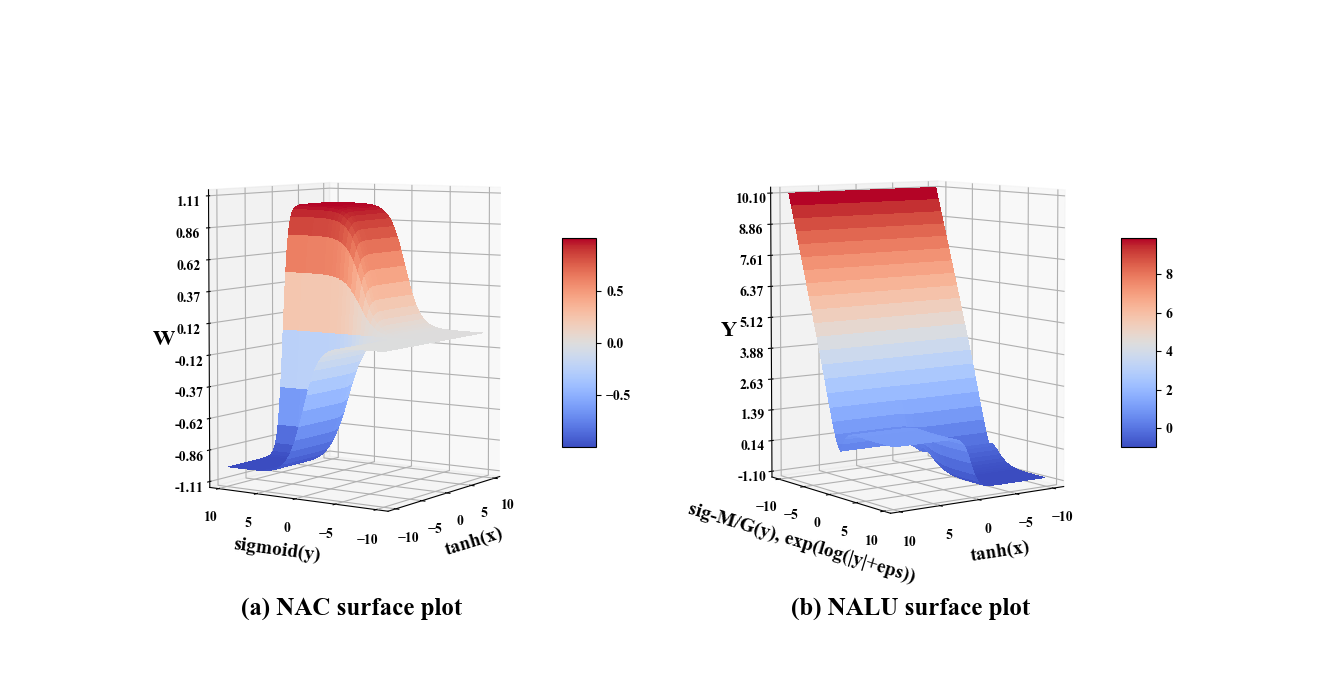}
\caption{
\textbf{Left:} NACs are biased towards learning [-1, 0, 1] as highlighted with large plateau regions around these values. This means its outputs are either addition or subtraction of input vectors not scaling.
\textbf{Right:} Approximate surface curve of NALU with some dimensional constraints for 3-D plotting. It highlights the ability to scale, along with earlier numerical biases around [-1, 0, 1] as shown with plateau region surfaces.}
\label{fig3}
\end{figure}

NACs \cite{b7} supports accumulation of numerical quantities additively, a desirable bias for linear exploration while counting. It is special type of linear layer with transformation matrix \textbf{W} being continuous and differentiable parameterization for gradient descent. W = $\tanh$($\hat{W}$)$\odot$$\sigma$($\hat{M}$)  consists of elements in [-1, 1] with bias close to  −1, 0, and 1. See figure \ref{fig3} for ideation of this concept with following equations for NAC:  a = Wx, W = $\tanh$($\hat{W}$)$\odot$$\sigma$($\hat{M}$) where \textit{$\hat{W}$}, \textit{$\hat{M}$} are learning parameters and \textit{W} is transformation matrix.

For complex mathematical operations like multiplication and division we use NALUs. It uses weighted sum of two sub-cells, one for addition or subtraction and another of multiply, division or power functions. It demonstrates that neural accumulators (NACs) can be extended for learning scaling operations with gate-controlled sub-operations. See figure \ref{fig3} for ideation of this concept with following equations for NALU: y = g$\odot$a + (1-g)$\odot$m; m = $\exp$W($\log$( $\mid$x$\mid$ + $\epsilon$)), g = $\sigma$(Gx) where \textit{m} is subcell that operates in \textit{log} space and \textit{g} is learned gate, both contains learning parameters.

\subsection{Comparative analysis of addition operation}

\begin{figure}[!h]
\centering
\includegraphics[width=0.50\textwidth]{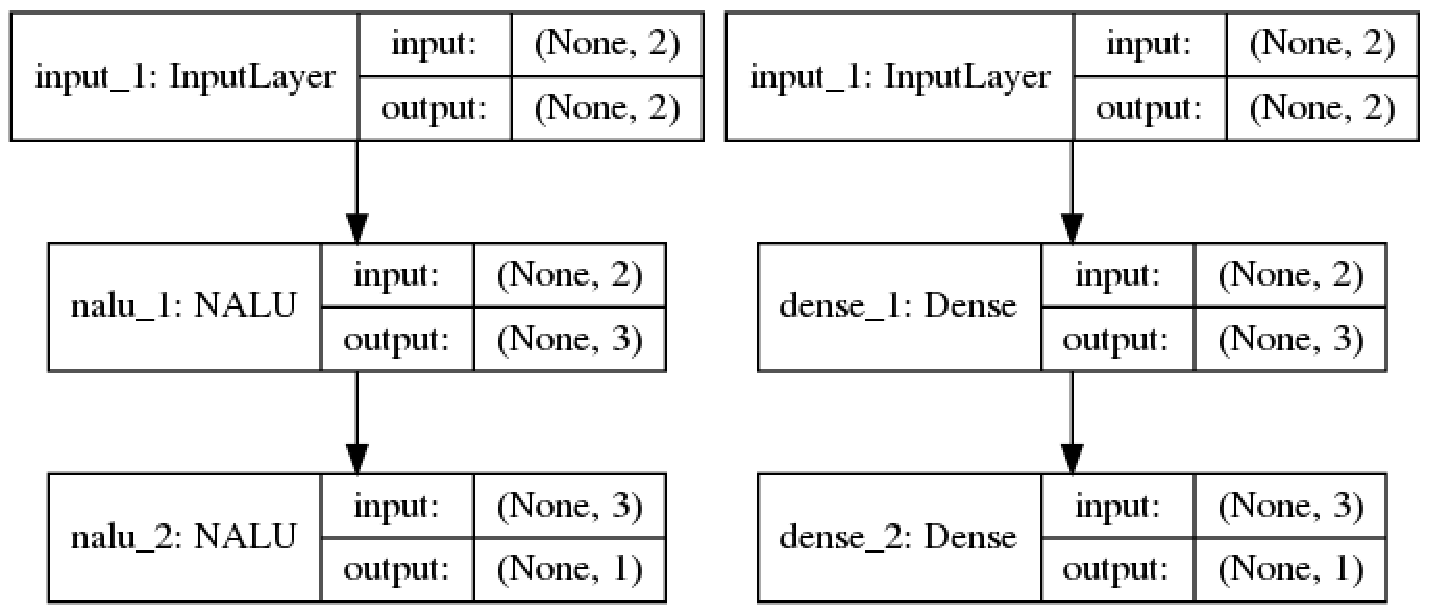}
\caption{The model on \textit{left} represents neural network with NAC or NALU units and on \textit{right}, a MLP which is trained with different activation functions. Each variant of above stated models have two inputs, three hidden units, one output.}
\label{fig4}
\end{figure}

Here, we use neural networks with NACs/NALUs and multilayer perceptrons (MLP) with different activation functions but same structures. These are trained with two randomly generated inputs from uniform distribution \textit{a} and \textit{b} with each having 2\textsuperscript{14} data points for training. Prediction capabilities on test data with values ranging up to 10 times the training range are evaluated as part of this experiment. Refer figure \ref{fig4} to observe architecture for both these trained models in this comparative study.

Comparative analysis is summarized in table  \ref{tab:table1} with MAE as an accuracy measure for MLP variants, NAC, NALU and its variants with changed learned gate for extrapolation. Also, in NALU-Tanh and NALU-Hard Sigmoid the learning gate \textit{g's} is changed to observe any improvements in NALU's performance based on this change.

\bgroup
\def\arraystretch{1.25}
\begin{table}[h!]
  \begin{center}
    \caption{\textbf{Comparative result summarization for multiple models}}
    \label{tab:table1}
    \begin{tabular}{|c|c|}
    
    \hline
      \textbf{Layer Configuration/Activations} & \textbf{Mean Absolute Error (\textit{a+b})} \\
    \hline

     Linear MLP & \num{3.63e-06}\\
     \hline
     Sigmoid MLP & \num{29.830}\\
     \hline
     Tanh MLP & \num{15.743}\\
     \hline
     ELu MLP & \num{0.019}\\
     \hline
     ReLU MLP & \num{0.001}\\
     \hline
     Leaky ReLU MLP & \num{9.83e-04}\\
     \hline
     PReLU MLP & \num{0.001}\\
     \hline
     NAC & \num{2.70e-06}\\
     \hline
     NALU & \num{2.71e-06}\\
     \hline
     NALU-Hard Sigmoid & \num{3.24e-06}\\
     \hline
     NALU-Tanh & \num{3.18e-06}\\
     \hline
    \end{tabular}
  \end{center}
\end{table}
\egroup

\begin{figure}[!h]
\centering
\includegraphics[width=0.50\textwidth]{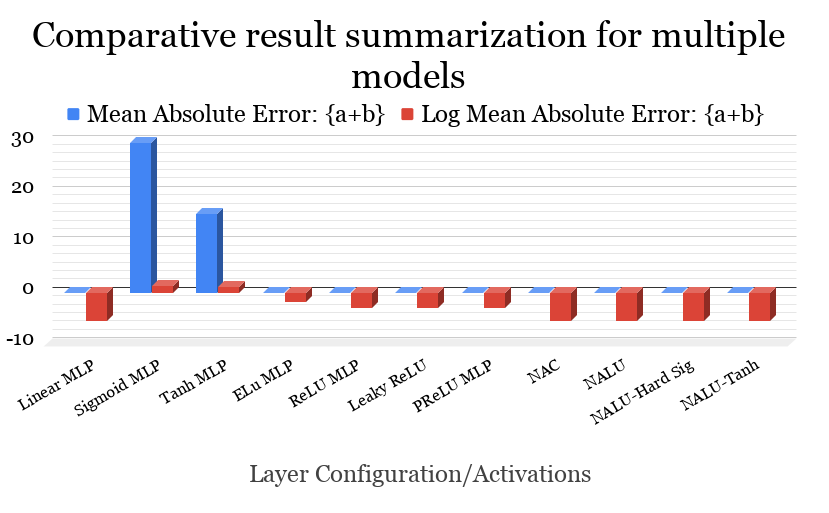}
\caption{Above visualization of MAE and log\textsubscript{10}(MAE) for different models learning identity mapping, demonstrates that our standard activation functions like \textit{sigmoid} and \textit{tanh} doesn't perform that well with higher data ranges during testing. Whereas, \textit{NACs} and \textit{NALUs} clearly overpowers the identity learning task and \textit{linear}, \textit{relu} based activation functions somewhat manages to provide acceptable results.}
\label{fig5}
\end{figure}

From these results stated in table \ref{tab:table1} and visualized in figure \ref{fig5} we conclude that Linear, LeakyReLU, ReLU activations and NAC, NALU, NALU-Tanh modules were the top performers in extrapolation task for numeric addition operation task. Hence, these top performers are used further in cell-counting task on synthetic dataset for learning end-to-end counting mechanism.

\subsection{Cell counting experiment}

\footnote{\textit{Code repository:} \url{https://github.com/ashishrana160796/nalu-cell-counting}}

In this experiment section the first subsections elaborates the datasets used for training and validation of our trained models, plus the data augmentation techniques used in our experiment. After that we elaborate onto different architectures used for training having different activation layers on standard architectures and residual concatenated connection modules on modified proposed model architectures.

\subsubsection{Datasets and data augmentation}

Synthetic dataset which is generated by system \cite{b29}. 200 highly-realistic synthetic fluorescence microscopic images of bacterial cells are used for experimentation with a 75/25 train-test split for training each model architecture and its variants. Images are having average of 174±64 cells.

\begin{figure}[!h]
\centering
\includegraphics[width=0.50\textwidth]{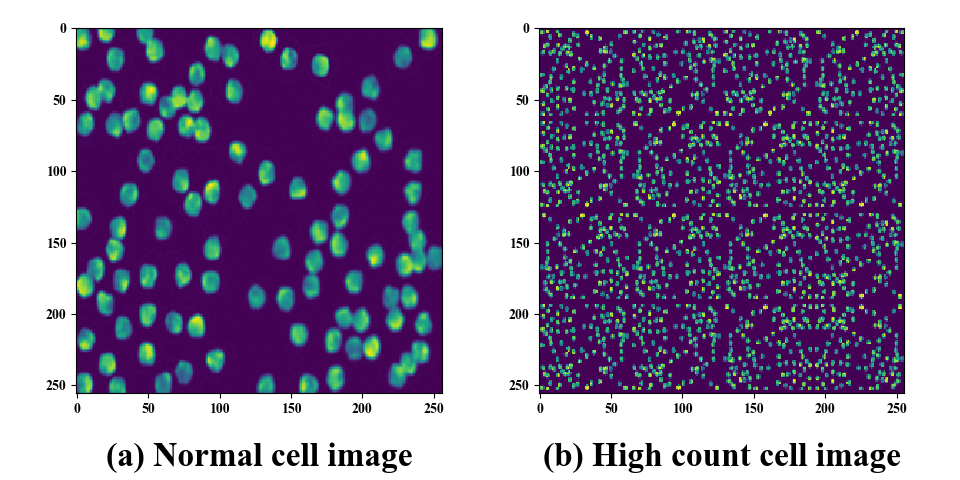}
\caption{\textit{Left: }Original image from BBBC005 dataset. \textit{Right: }Repeated image generated from a \textit{4X4} grid repetition operation with random horizontal and vertical flips of base image shown in left.}
\label{fig6}
\end{figure}

For validation of trained models and checking true generalization capabilities we use BBBC005 from the the Broad Institute’s Bioimage Benchmark Collection \cite{b10}. This dataset is comprised of 600 images have a corresponding foreground mask which alters the focus on these images and ground truth images are completely in-focused version before any guassian filter application. We take a subset of this dataset with highly focused \textit{F1} images only and their corresponding ground truth image for validating our model. And after that we coalesce each image 16 times in 4x4 grid manner with random vertical and horizontal flips to create a final high cell count images for our validation dataset, see figure \ref{fig6}. The ground truth images are accordingly changed with same randomness and they are eroded also to match the fluorescent synthetic dataset on which the models are trained on.

\begin{figure}[!h]
\centering
\includegraphics[width=0.50\textwidth]{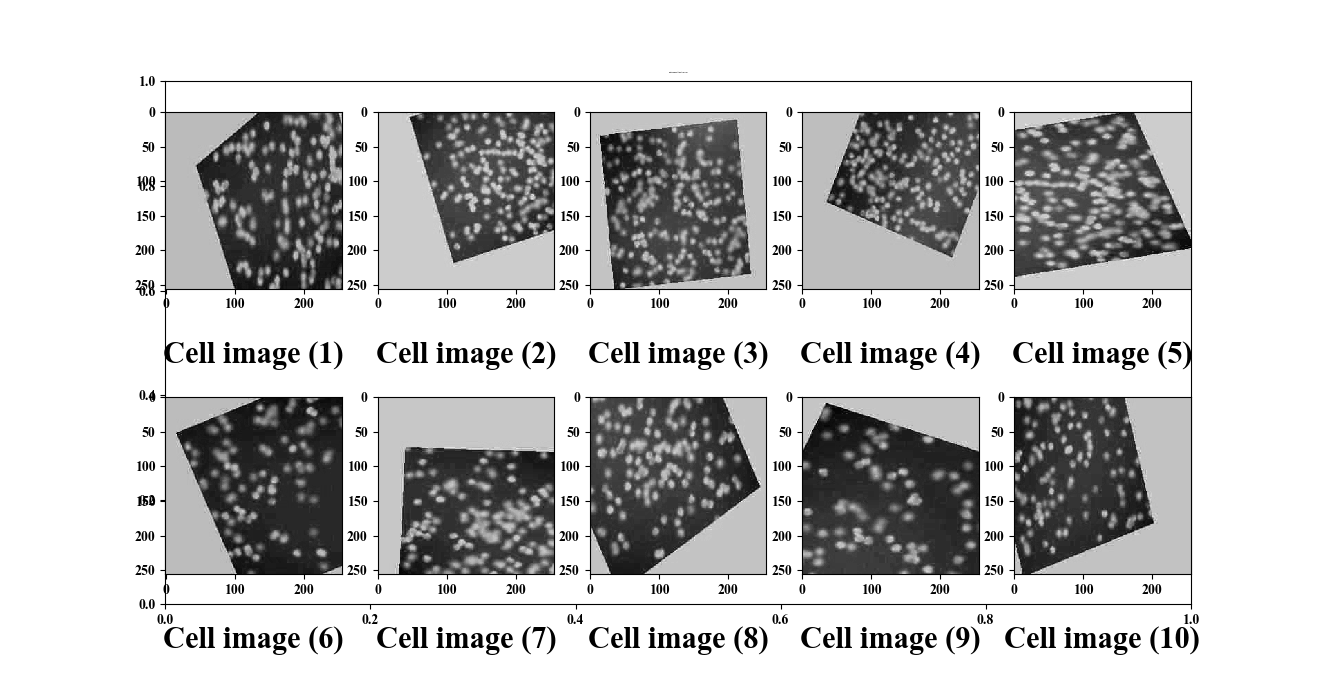}
\caption{The above figure illustrates different cell augmentation techniques in action like elastic deformations, random rotations and plane shifting for the input images adding more robustness to the trained model.}
\label{fig7}
\end{figure}

Data augmentation with elastic deformations to training images is applied for teaching network the desired invariance and robustness properties, like specified in figure \ref{fig7}. These elastic deformations are introduced in the form of angular shear in the training images. Translation and rotation invariance along with robustness to gray value variations and deformations is main focus of augmentation process for microscopic images. Disfigurement using random displacement vectors on a coarse 3x3 grid are also generated. These data augmentation techniques especially are helpful for our custom data which is created just by repeating the original image in order to supplement a more robust dataset for the model to train on.

\subsubsection{Defining regression task and architecture details}

In training dataset ground truth is provided as dot annotation corresponding to each cell image. For training, dot annotations are represented by Gaussian and density surface D(x) which is formed from superposition of Gaussians. The optimization task is to regress density surface from corresponding image I(x). This is achieved by training convolutional neural networks (CNN) using mean square error between output heat map and target density surface as the loss function. Hence, at inference given an input I(x), the model predicts density heat map D(x).

\begin{figure}[!h]
\centering
\includegraphics[width=0.45\textwidth]{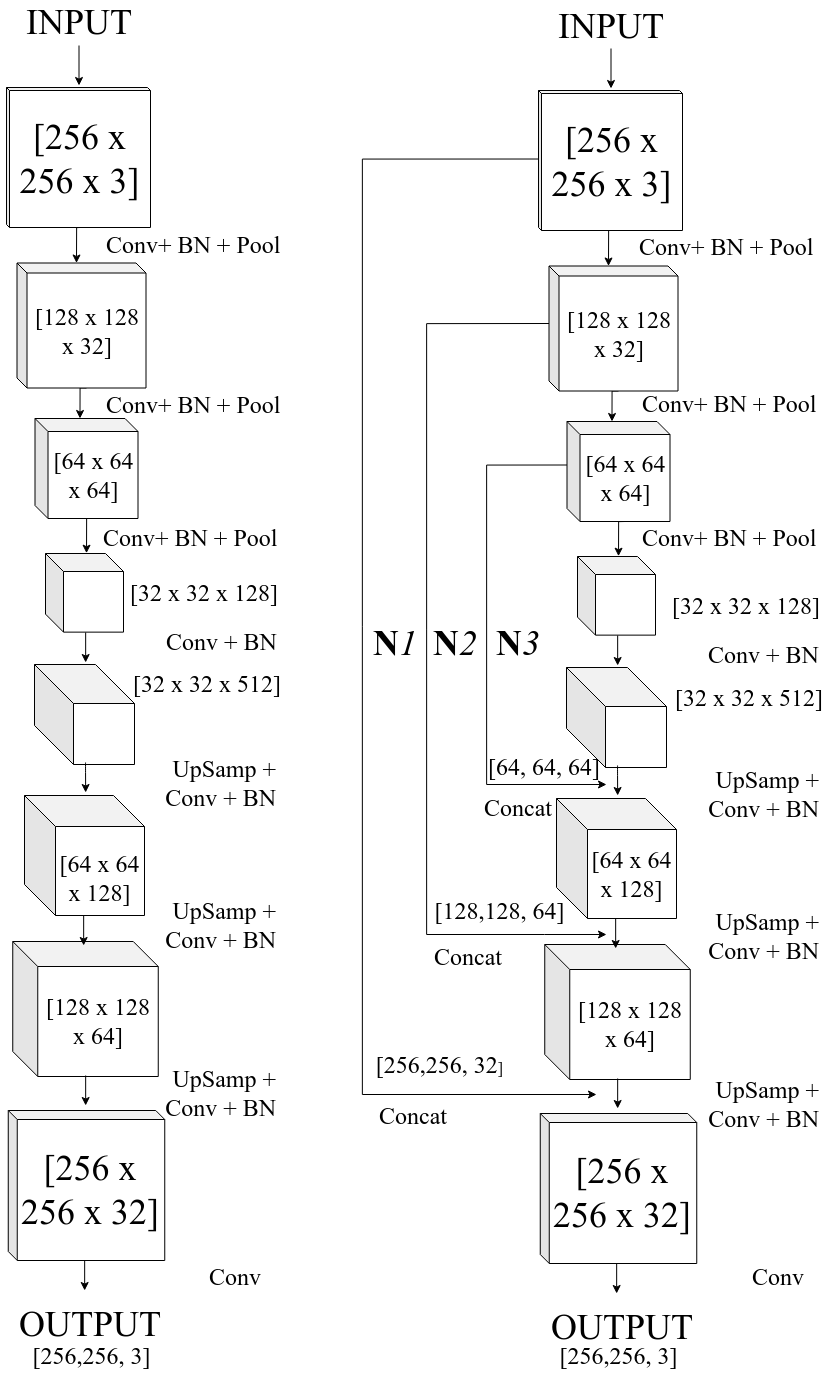}
\caption{\textit{Network Architecture Left: }FCRN with 3x3 convolution operations. \textit{Network Architecture Right: } FCRN added with residual concatenation connections of \textbf{N}\textit{i} numerically biased units. Also, for dimensional compatibility \textbf{N}\textit{i} residual layers are fed to corresponding main network layers and after that normalized for regularization. At last, 1x1 \textbf{Conv} operation output in the form of density map of result image is generated .\newline
\textbf{Conv + BN + Pool: }A 3x3 convolutional operation with batch normalization regularization and  2x2 max pooling layer.\newline
\textbf{Unsample+Conv + BN: }Unsampling the image and then applying convolutional operation with batch normalization regularization.\newline
\textbf{Concat: }Feature maps concatenated along depth dimension.\newline
\textbf{N\textit{i}: }NAC or Variants of NALUs as residual concatenated connection.
}
\label{fig8}
\end{figure}

FCRNs are inspired from VGG-net, we only used small kernels of size 3x3 pixels for designing our network. Feature maps are increased for avoiding spatial information loss. Activation layers like convolution-ReLU-Pooling are popular in CNN architectures \cite{b24}. Here, we have altered these layers to create different models which contains some numerical bias in the form of residual connections and regularized by batch normalization. The first layers contains convolutions-pooling operations, then we undo spatial reduction by upsampling operations for end-to-end model training. Also, for dimensional compatibility of residual NAC or NALU modules we append their output to corresponding main network layer across depth dimension and batch normalize the output after appropriate convolutional operation. See figure \ref{fig8} for comparison between earlier original model and newly proposed architecture along with architecture parameter details.

\begin{figure}[!h]
\centering
\includegraphics[width=0.40\textwidth]{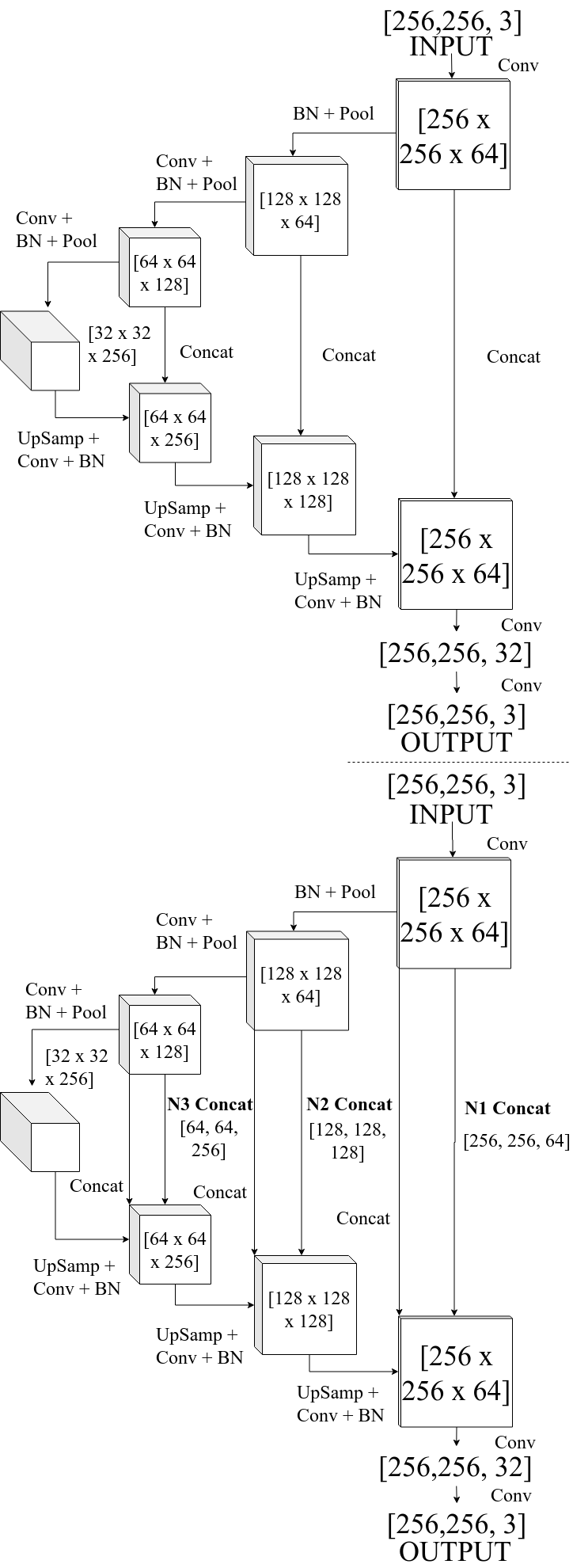}
\caption{\textit{Network Architecture Top: }U-net with 3x3 convolution operations and increasing depth of filters from 64 to 256 for feature abstraction \& learning and after that these layers are fed to upsampling layers. \textit{Network Architecture Bottom: } U-nets main network added with residual concatenation connections of \textbf{N}\textit{i} numerically biased units and after that batch normalized for regularization. Similar, operation abbreviations used as stated in figure \ref{fig8} \newline}
\label{fig9}
\end{figure}

U-net is modification upon the previously discussed FCRN architecture by having large number of feature channels for upsampling to propagate context information to high resolution layers. That makes expansive path almost symmetric to contracting path yielding a u-shape. Similar to above FCRN's optimization problem formulation the evaluation parameters remains the same. Residual concatenated connection addition with NACs and NALU units along with batch normalization is done for adding numerical bias information to the main network. Also, U-net architecture used in this paper is more computationally expensive than FCRN having approximately thrice the number of parameters leading to more feature learning capacity and is expected to perform better for prediction tasks. See figure \ref{fig9} for comparison between earlier original U-net model and newly proposed architecture along with parameter details.

For the concatenation of residual connections of these units the dimensional consistency is maintained by added them along the depth dimension with the main network. After that upsampling, convolutional and batch normalization regularization operations are applied accordingly for these units to merge with the base network. FCRN’s implementation resembles that of MatConvNet \cite{b30} as upsampling in Keras is implemented by repeating elements, instead of bilinear sampling. In U-nets, low-level feature representations are fused during upsampling, aiming to compensate the information loss due to max pooling.

\section{Results}

MAE is the metric used in this paper for measuring results for cell counting on the synthetic cell dataset \cite{b29} and custom BBBC005 synthetic modified high cell count validation dataset.

\begin{itemize}
\item \textbf{Mean Absolute Error (MAE):}
Mean Absolute Error (MAE): The mean absolute error is an average of the difference between the predicted value and true value.
\begin{equation}
AE = \|e_i\| = \|y_i - x_i\|
\end{equation}
\begin{equation}
MAE = \sum_{i=1}^{n} \frac{|e_i|}{n}
\end{equation}
\end{itemize}

\begin{itemize}
\item \textbf{Relative Improvement  Percentage}
Relative Improvement  Percentage (RIP): Here, in context of this paper it defined as percentage improvement in MAE of a given model with respect to baseline ReLU models for FCRN and U-net architectures. In below equation, M\textsubscript{r} is MAE from baseline ReLU model and M\textsubscript{i} is model under consideration.
\begin{equation}
 RIP_\% = ((M_r - M_i) / M_r) * 100
\end{equation}

\end{itemize}

Result table \ref{tab:table2} compares baseline FCRN, U-net architectures with new numerically biased ResNet like  connection modules with NACs and NALUs units under current training setup. With our setup we able to obtain similar results as mentioned in earlier reference papers and also we have equipped earlier model architectures with different regularization activations as specified in the table. From earlier ReLU implementation clearly Linear and LeakyReLU activation regularization based baseline models have performed well. Also, for both model structures NAC and NALUs residual modules have outperformed all the earlier specified regular FCRN architecture. And similar results are extended by U-net model evaluation where NALU layer concatenation based U-net outperforms all the models trained for our experiment.

\bgroup
\def\arraystretch{1.25}
\begin{table}[h!]
  \begin{center}
    \caption{\textbf{Result summarization for trained models}}
    \label{tab:table2}
    \begin{tabular}{|c|c|c|c|}
    
    \hline
      \textbf{FCRN-Models} & \textbf{MAE} & \textbf{U-Net-Models} & \textbf{MAE} \\
    \hline

     ReLU  & 3.43 & ReLU & 1.78\\
     LeakyReLU  & 3.39 & LeakyReLU & 1.74\\
     Linear  & 3.34 & Linear & 1.73\\
	 NALU-tanh & 3.21 & NALU/tanh & 1.56\\
	 NALU & \textbf{3.17} & NALU & \textbf{1.42}\\
     NAC & 3.23 & NAC & 1.63\\
     
     \hline
     \end{tabular}
  \end{center}
\end{table}
\egroup

Result table \ref{tab:table3} compares performance of above trained models on a new validation dataset containing much higher cell counts for measuring performance on extrapolation capabilities counting tasks. For validation set we have used 300 images of size 256x256 pixels with cell counts averaging around 1200±12. Here also, NAC and NALU based residual concatenation module based models outperforms earlier architectures for counting tasks. But, this time the relative improvement is even more for NALU based FCRN and U-net models showcasing better generalization abilities of trained models.

\bgroup
\def\arraystretch{1.25}
\begin{table}[h!]
  \begin{center}
    \caption{\textbf{Validating trained models for extrapolation cell counting tasks}}
    \label{tab:table3}
    \begin{tabular}{|c|c|c|c|}
     
         \hline
      \textbf{FCRN-Models} & \textbf{MAE} & \textbf{U-Net-Models} & \textbf{MAE} \\
    \hline

     ReLU  & 3.04 & ReLU & 2.87\\
     LeakyReLU  & 2.99 & LeakyReLU & 2.62\\
     Linear  & 2.85 & Linear & 2.47\\
	 NALU-tanh & 2.32 & NALU-tanh & 1.95\\
	 NALU & \textbf{2.27} & NALU & \textbf{1.87}\\
     NAC & 2.40 & NAC & 1.92\\
    
     \hline
     \end{tabular}
  \end{center}
\end{table}
\egroup

\begin{figure}[!h]
\centering
\includegraphics[width=0.50\textwidth]{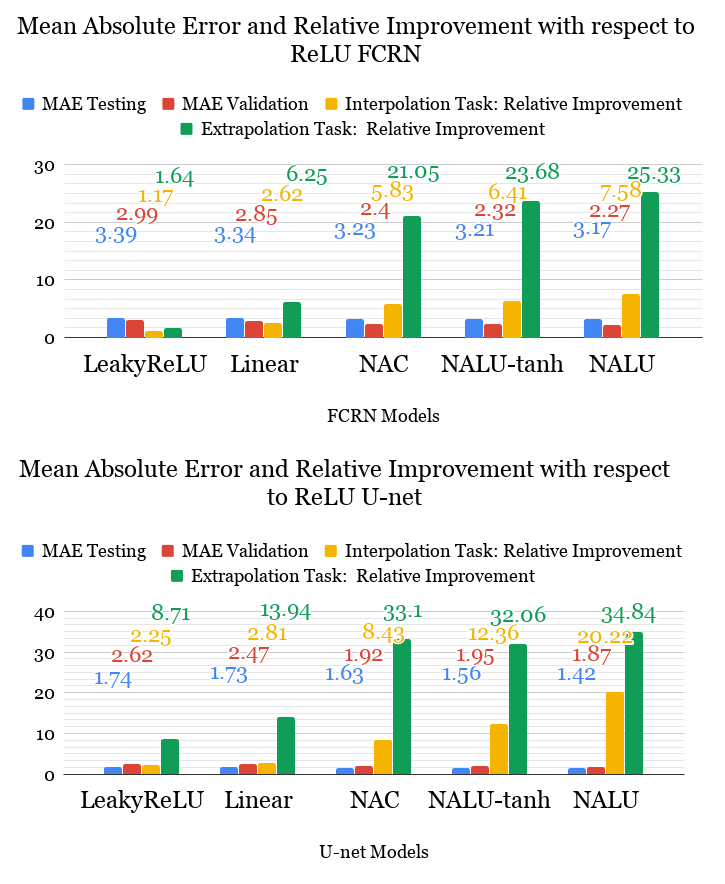}
\caption{Moving towards right in above figure and measuring RIP metric there is sharp increase with NALU \& NAC units based models demonstrating even more increase in relative improvement with respect to baseline ReLU models for extrapolation tasks. For interpolation task there is highest 9\% improvement and for extrapolation there is 23\% improvement with NAC units as residual connection.}
\label{fig10}
\end{figure}

Relative improvement in predictions is visualized in figure \ref{fig10} against ReLU based activation base result for comparison with other activation layer changes in FCRNs/U-nets and concatenation layer NALU/NAC residual connection addition in FCRNs \& U-nets. It includes averaged out comparison from multiple executions of training and testing runs for both interpolation testing and extrapolation validation counting tasks for FCRN and U-net variant models with respect to ReLU based FCRN and U-net model. From, this figure it is clearly highlighted that models with NAC and NALUs residual modules have better generalization capabilities for extrapolation counting tasks i.e. they are better generalizers for this given cell counting task with increase in relative improvement in prediction as compared to base ReLU implementation. Here, for measuring generalization capabilities of model in extrapolation task relative improvement metric is selected as it bring more perspective to the improved performance on validation dataset. The performance has shown a general increment for every model as validation dataset consist of highly focused images. Hence, measuring relative improvement in results is more appropriate decision to get understanding of true improvment in model prediction capabilities.

This figure \ref{fig10} shows more increase in relative improvement as we move right towards horizontal axis for both testing and validation task with extrapolation where in validation extrapolation task NAC/NALU models from which we can conclude that trained models are having better generalization abilities with some learned numerical bias in their trained weights with which even better predictions for higher count cells is made.

\section{Summary}
We were able to demonstrate that addition of newly proposed NAC and NALU units in existing architectures in the form of residual concatenation connection layer modules achieves better results. With numerically biased residual connections, higher accuracy for more dense images having higher counts of cells is achieved. Hence, producing more generalized cell counters that provides better predictions for real life use-cases. Finally, for code implementation details and other supplementary experimental results refer to this paper's \href{https://github.com/ashishrana160796/nalu-cell-counting}{github repository}.

\end{document}